%% file: acl_latex.tex
\documentclass[11pt]{article}

\usepackage[final]{acl}

\usepackage{times}
\usepackage{latexsym}

\usepackage[T1]{fontenc}

\usepackage[utf8]{inputenc}

\usepackage{microtype}

\usepackage{inconsolata}

\usepackage{graphicx}

\usepackage{algorithm}
\usepackage{algpseudocode}
\usepackage{amsmath}
\usepackage{booktabs}   
\usepackage{multirow}   
\usepackage{amssymb}    
\usepackage{makecell}
\usepackage[table]{xcolor}
\usepackage[dvipsnames]{xcolor}
\definecolor{upgreen}{rgb}{0.0, 0.5, 0.0}
\definecolor{lightblue}{rgb}{0.9, 0.95, 1}
\usepackage{tabularx}   
\usepackage{enumitem}   

\usepackage[most]{tcolorbox}
\usepackage{listings}

\definecolor{promptgray}{RGB}{245, 245, 245}
\definecolor{commentgreen}{RGB}{0, 128, 0}
\definecolor{keywordpurple}{RGB}{128, 0, 128}

\lstdefinelanguage{json}{
    basicstyle=\small\ttfamily,
    numbers=none,
    breaklines=true,
    frame=none,
    stringstyle=\color{blue},
    keywordstyle=\color{keywordpurple},
}

\newtcolorbox{PromptBox}[1]{
    colback=gray!5,
    colframe=gray!75,
    fonttitle=\bfseries\sffamily,
    title=#1,
    arc=1mm,
    outer arc=1mm,
    left=5pt,
    right=5pt,
    top=5pt,
    bottom=5pt,
    breakable 
}

\title{CT-Flow: Orchestrating CT Interpretation Workflow with Model Context Protocol Servers}

\author{
  \textbf{Yannian Gu}$^{1*}$, 
  \textbf{Xizhuo Zhang}$^{1*}$, 
  \textbf{Linjie Mu}$^{1}$, 
  \textbf{Yongrui Yu}$^{1}$, 
  \textbf{Zhongzhen Huang}$^{1}$, \\
  \textbf{Shaoting Zhang}$^{3\dagger}$, 
  \textbf{Xiaofan Zhang}$^{12\dagger}$ \\
  \\
  $^1$Qing Yuan Research Institute, Shanghai Jiao Tong University, Shanghai, China \\
  $^2$Shanghai Innovation Institute, Shanghai, China \\
  $^3$Sensetime Research, Shanghai, China \\
  \small{$^*$Equal contribution, $^\dagger$Corresponding authors}
}

\begin{document}
\maketitle

\input{sections/0_abstract}
\input{sections/1_introduction}

\input{sections/2_related_work}
\input{sections/3_proposed_method}
\input{sections/4_dataset}
\input{sections/5_experiments}
\input{sections/6_conclusion}
\input{sections/7_limitation}
\input{sections/8_ethical}

\bibliography{custom}

\appendix
\input{sections/10_appendix}

\end{document}

%% file: sections/0_abstract.tex
\begin{abstract}
Recent advances in Large Vision–Language Models (LVLMs) have shown strong potential for multi-modal radiological reasoning, particularly in tasks like diagnostic visual question answering (VQA) and radiology report generation.
However, most existing approaches for 3D CT analysis largely rely on static, single-pass inference.
In practice, clinical interpretation is a dynamic, tool-mediated workflow where radiologists iteratively review slices and use measurement, radiomics, and segmentation tools to refine findings.
To bridge this gap, we propose CT-Flow, an agentic framework designed for interoperable volumetric interpretation.
By leveraging the Model Context Protocol (MCP), CT-Flow shifts from closed-box inference to an open, tool-aware paradigm.
We curate CT-FlowBench, the first large-scale instruction-tuning benchmark tailored for 3D CT tool-use and multi-step reasoning.
Built upon this, CT-Flow functions as a clinical orchestrator capable of decomposing complex natural language queries into automated tool-use sequences.
Experimental evaluations on CT-FlowBench and standard 3D VQA datasets demonstrate that CT-Flow achieves state-of-the-art performance, surpassing baseline models by 41\% in diagnostic accuracy and achieving a 95\% success rate in autonomous tool invocation. 
This work provides a scalable foundation for integrating autonomous, agentic intelligence into real-world clinical radiology.
\end{abstract}

%% file: sections/1_introduction.tex
\section{Introduction}
\label{sec:Introduction}

Computed Tomography (CT) is a cornerstone of modern diagnostic radiology~\cite{al2025trends}. The subtle 3D radiographic patterns (e.g., small hemorrhages, early ischemic changes, faint ground-glass opacities) and measurement-dependent criteria (e.g., size thresholds, volumetric assessments, and attenuation values in Hounsfield units) often critically influence downstream triage and treatment decisions~\cite{wang2024enhancing,mao2025ct}. The rapidly increasing volume and complexity of CT examinations have consequently fueled strong interest in automated and assistive interpretation systems. Recent advances in Large Vision–Language Models (LVLMs) offer a promising avenue to alleviate physician workload~\cite{li2023llava,tu2024towards,jiang2025hulu}. By learning from large-scale paired imaging and clinical text data, these models have demonstrated notable capabilities in diagnostic visual question answering (VQA)~\cite{zhang2023pmc} and automated report generation~\cite{hamamci2024ct2rep,zhao2024ratescore}, highlighting their potential role in radiographic interpretation.

\begin{figure}
    \centering
    \includegraphics[width=1\linewidth]{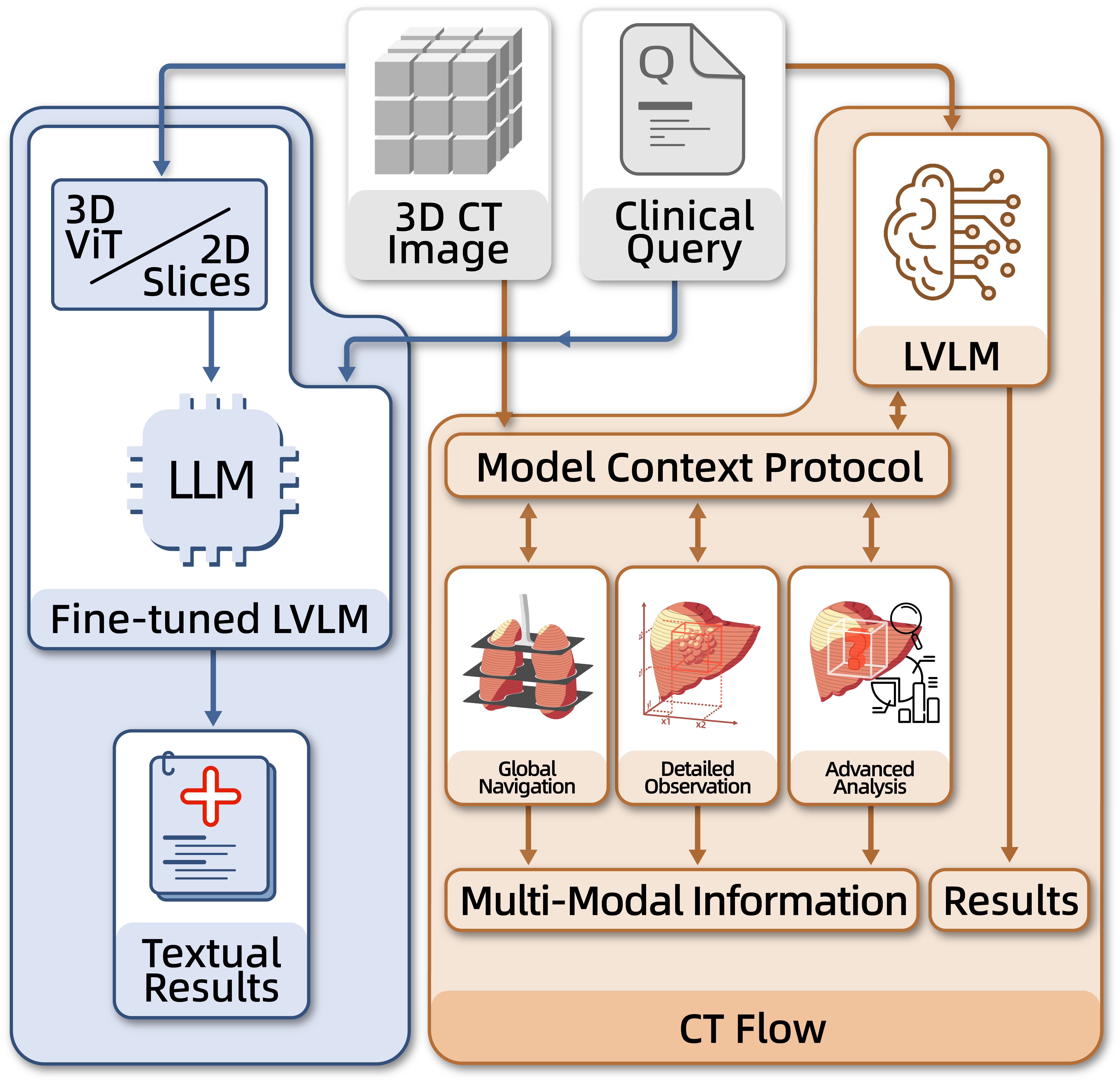}
    \caption{Comparison of 3D CT analysis paradigms. Left: Traditional End-to-End LVLMs rely on passive visual ingestion of 3D data, resulting in static textual outputs. Right: The proposed CT-Flow framework leverages the Model Context Protocol to transform the LLM into an active agent. It dynamically orchestrates specialized tools to deliver precise, multi-modal diagnosis.}
    \label{fig:placeholder}
\end{figure}

Despite these advances, current LVLM-based approaches remain poorly aligned with real-world CT interpretation workflows. Most existing methods treat CT volumes as static visual inputs and rely on end-to-end inference over pre-encoded representations. To handle 3D data, prior work either employs specialized 3D encoders (e.g., 3D Vision Transformers) to aggregate voxel-level features~\cite{bai2024m3d} or adapts 2D encoders to process serialized axial slices~\cite{hamamci2024developing}. Although effective for capturing global context, these strategies inevitably introduce information bottlenecks~\cite{wu2025towards} that obscure fine-grained anatomical details and subtle radiographic cues that clinicians rely on for evidence-based decision making~\cite{wu2025vision}. More fundamentally, clinicians rarely arrive at diagnoses through a single passive observation. Instead, CT interpretation is an inherently active and iterative process involving scrolling through slices, switching planes, probing voxel densities, measuring lesions, and invoking specialized tools for segmentation or radiomics analysis~\cite{ritchie2025impact}. Yet most existing LVLMs operate in a ``read-only'' mode, lacking the agency required for iterative verification and hypothesis refinement~\cite{goswami2025medivlm,friebe2025ai}. To better reflect clinical reality, we argue that CT interpretation should be reframed as an agentic problem rather than a purely perceptual one.

Recent progress in large language models (LLMs) has demonstrated that equipping models with explicit tool access can substantially improve problem-solving capability and response reliability. The recently proposed Model Context Protocol (MCP)~\cite{anthropic2024mcp} further advances this paradigm by providing a standardized interface that connects LLMs to external data sources and tools, reducing reliance on bespoke integrations. The rapid adoption of MCP in general software ecosystems suggests a promising direction for transforming LVLMs from static predictors into dynamic orchestrators capable of interfacing with clinical utilities in a scalable and standardized manner.

In this work, we introduce CT-Flow, the first agentic framework that incorporates MCP to transform passive volumetric encoding into active, tool-mediated probing. CT-Flow integrates four MCP servers supporting navigation, measurement, segmentation, and radiomics analysis. Within this framework, 3D CT understanding is formulated not as a single perception task but as a sequential decision-making process. Given a diagnostic query, the model dynamically decomposes the task into a series of tool calls, such as multi-planar visualization, region-specific segmentation, and quantitative radiomics extraction, and grounds its reasoning in tool-verified evidence. This design mitigates the information bottlenecks inherent to end-to-end 3D LVLMs and aligns model behavior more closely with real clinical workflows.

To support this paradigm, we introduce CT-FlowBench, a benchmark designed to evaluate agent trajectories in 3D CT clinical workflows. Unlike existing static VQA datasets that focus primarily on final-answer correctness, CT-FlowBench formalizes CT interpretation as executable reasoning chains and provides supervision over intermediate decisions and tool usage. Building upon CT-Flow, we establish a human–AI collaborative pipeline in which experts and agents iteratively co-design and refine diagnostic queries along with their ground-truth trajectories. This process yields 2,000 samples for training and 300 samples for evaluation.

We conduct extensive experiments on established 3D CT VQA benchmarks as well as CT-FlowBench. LVLMs equipped with CT-Flow achieve substantial performance gains and outperform specialized medical LVLMs, demonstrating the effectiveness of reframing CT understanding as agentic reasoning with tool use. Moreover, models fine-tuned on CT-FlowBench-train exhibit strong generalization and robustness, indicating a promising new paradigm for developing clinically aligned AI systems for CT interpretation. 

We summarize our contributions as follows:

\begin{itemize}[leftmargin=*, noitemsep]
 \item We propose \textbf{CT-Flow}, a novel agentic architecture that leverages the MCP to shift 3D medical analysis from passive encoding to active, tool-mediated probing, aligning model behavior with clinical workflows.
 \item We introduce \textbf{CT-FlowBench}, the first benchmark dedicated to training and evaluating medical agents on 3D CT workflows, providing a standardized testbed for agentic reasoning in radiology.
 \item We demonstrate that CT-Flow yields substantial performance improvements while producing transparent, traceable, and clinically aligned reasoning processes compared to end-to-end LVLM baselines.
 \end{itemize}

%% file: sections/2_related_work.tex
\begin{figure*}[h]
    \centering
    \includegraphics[width=1\linewidth]{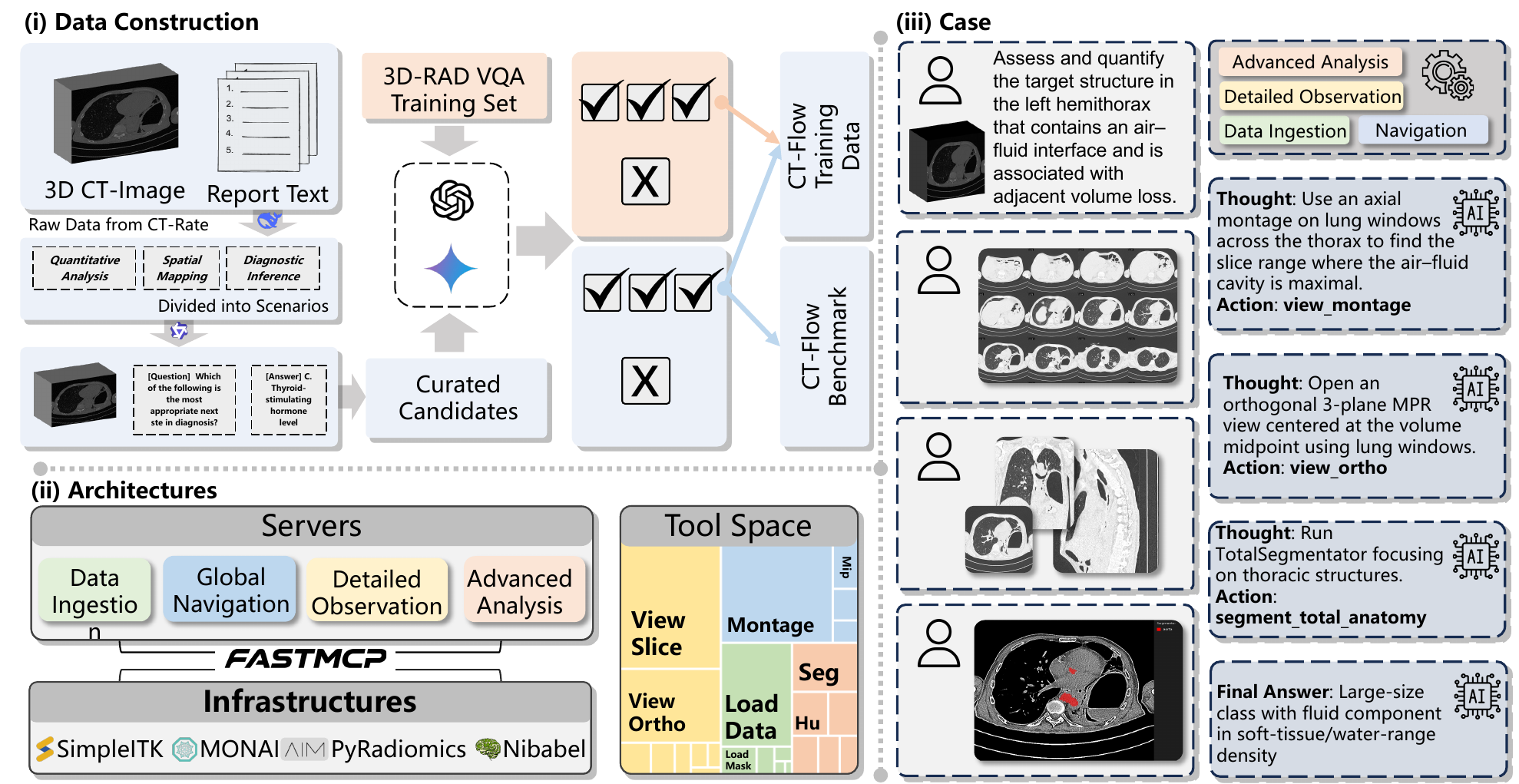}
    \caption{Overview of the CT-Flow framework. (i) Data Construction: The pipeline for raw data curation, trajectory synthesis, and the establishment of the CT-Flow benchmark.
(ii) Architectures: The system decouples the LLM orchestrator from the imaging environment via FASTMCP, bridging high-level servers with medical imaging infrastructures to provide a suite of atomic tools in the Tool Space.
(iii) Case Study: A demonstration of a Language-Action Trajectory $\mathcal{T}$. The orchestrator performs Active Probing by iteratively generating reasoning states ($s_t$), executing tool calls ($a_t$), and interpreting high-fidelity observations ($o_t$) to reach a grounded diagnostic answer.}
    \label{fig:pipeline}
\end{figure*}

\section{Related Work}
\label{sec:Related Work}

\paragraph{3D Volumetric Interpretation in Medical VLMs.}
Research on 3D medical vision-language modeling generally follows two paradigms for volumetric interpretation.
\textit{Native 3D approaches} directly encode volumes with 3D backbones: RadFM \cite{wu2025towards} unifies 2D and 3D images via 3D ViT. M3D \cite{bai2024m3d} deals with volumetric data via 3D spatial pooling. However, they suffer from high computational cost and limited fine-grained detail retention~\cite{ates2025dcformer}.
\textit{Serialization-based methods} treat CT volumes as long slice sequences: OmniCT \cite{anonymous2025omnict} aggregates cross-slice context via sequence modeling but may lose spatial fidelity. Hulu-Med \cite{jiang2025hulu} unifies 2D and 3D inputs within a transparent generalist medical VLM by decomposing 3D volumes into constituent slices. Vote-MI \cite{wang2024enhancing} selects representative slices to reduce computation, at the risk of discarding diagnostically critical spatial cues. While these methods excel at capturing global patterns, their reliance on static, lossy encoding may hinder the retention of fine-grained details necessary for precise diagnosis~\cite{zhong2025vision}.

\paragraph{Agentic Reasoning and Tool Orchestration.}
Leveraging an LLM as the reasoning backbone, an autonomous agent should interact with its environment to make decisions and take actions.
ReAct~\cite{yao2022react} serves as a general agent paradigm that combines reasoning and acting in a unified framework, which enables LLMs to conduct reasoning for making plans and taking actions, such as tool invocation, and to incorporate information from the environment.
To facilitate LLMs in invoking orchestrated tools for solving complex tasks, the MCP~\cite{anthropic2024mcp} provides an open and standardized protocol for managing and executing tools.
For example, AgentMaster~\cite{liao2025agentmaster} utilizes MCP to provide a unified interface for tool access, long-term memory, and context management for multi-modal information retrieval. MCP facilitates interactions between LLM reasoning and specialized tool invocation.

\paragraph{Medical Agents and Clinical Tool-use.}
Using Large Language Models (LLMs) as controllers for external tools has gained significant traction in medical diagnostics.
Early systems like ChatCAD \cite{wang2024interactive} and ChatCAD+ \cite{zhao2024chatcad+} pioneered the use of LLMs to integrate outputs from various computer-aided diagnosis tools into interactive reports.
Recent research, such as Med-Agents \cite{tang2024medagents}, explores multi-agent frameworks to facilitate collaborative clinical reasoning across different specialized domains.
MedRAX \cite{fallahpour2025medrax} demonstrates the potential of LLMs in performing strategic planning and complex reasoning for radiology-specific tasks.
Other efforts focus on integrating visual grounding utilities (e.g., localized segmentation and detection) into LLM's reasoning loop.
These agentic frameworks typically aim to transform LLMs from simple text generators to clinical co-pilots capable of orchestrating diverse tools. However, many of them treat tool invocation as isolated or fragmented actions, which fails to capture the iterative nature of complex 3D analysis.

%% file: sections/3_proposed_method.tex
\section{Methodology}
\label{sec:Methodology}

CT-Flow aims to transform LVLMs from static predictors into clinical workflow orchestrators, enabling them to make diagnostic plans, invoke diverse imaging tools on demand, and complete CT tasks through iterative verification. An overview of CT-Flow is illustrated in Fig.~\ref{fig:pipeline}.

\subsection{Standardizing Clinical Interface via MCP}
\label{sec:mcp_interface}

To enable a model to reason across complex 3D volumes, we first define a standardized action space by abstracting heterogeneous imaging operations into a composable toolchain via the MCP. In practice, we consolidate high-frequency, clinically essential capabilities in CT workflows into four tool suites, spanning the full pipeline from data loading to pre-decision verification:

\begin{itemize}[leftmargin=*, noitemsep]
\item \textit{Data Ingestion}: Ingests CT volumes and metadata into a standardized, queryable 3D state for downstream tool execution.

\item \textit{Global Navigation}: Enables fast whole-volume orientation and coarse anatomical localization to guide subsequent probing.

\item \textit{Detailed Observation}: Retrieves targeted high-resolution views (slices or sub-volumes) to verify diagnostic hypotheses with local evidence.

\item \textit{Advanced Analysis}: Provides quantitative and structured measurements (e.g., Hounsfield Units or segmentation) to support decisions.
\end{itemize}

These tool suites collectively form the atomic action space. By encapsulating low-level image processing within the MCP framework, we provide the necessary infrastructure for the model to perform iterative, goal-oriented probing, shifting the paradigm from static, single-step prediction to dynamic orchestration of clinical workflows.

\subsection{Iterative Probing over ReAct}
\label{sec:react_trajectory}

Building upon the standardized action space defined above, we formalize the diagnostic process as a Reasoning-Acting Trajectory. Following the ReAct paradigm, the orchestrator drives the clinical workflow by interleaving its internal ratiocination with active environmental probing. For a given clinical query $Q$, the system generates a sequential execution path:
\begin{equation*}
\mathcal{T} = \{(s_0, a_0, o_0), (s_1, a_1, o_1), \dots, (s_n, a_n, o_n)\}
\end{equation*}
where $s_t$ denotes the reasoning state (the model's thought process for interpreting findings and planning the next step), $a_t \in \mathcal{A}$ represents a specific action issued via the MCP interface, and $o_t$ is the observation (e.g., visual evidence or quantitative metrics) returned by the imaging environment.

The significance of this trajectory-based formulation is that it transforms the diagnostic task into an active information-probing process. 
Instead of relying on a single-pass prediction, the orchestrator iteratively refines its understanding of the case. 
If a particular observation $o_t$ provides insufficient information to resolve the query, the model leverages its updated reasoning state $s_{t+1}$ to adjust its hypothesis and initiate further targeted probing. 
This iterative loop ensures that the final diagnostic answer $A$ is synthesized from the accumulated evidence collected throughout the trajectory, allowing the model to handle complex clinical cases that require multi-step examination.

    
    

%% file: sections/4_dataset.tex
\section{Dataset Construction}
\label{sec:data_construction}

Here we describe the construction of CT-FlowBench, a benchmark that represents 3D chest CT cases as \emph{executable} reasoning trajectories.
\paragraph{Design Rationale.}
Clinical volumetric interpretation is inherently interactive: radiologists navigate across slices, localize regions of interest, perform measurements, and iteratively integrate evidence.
To reflect this workflow and enable step-level verification, we construct CT-FlowBench with tool-mediated trajectories whose intermediate observations are execution-retrievable from the raw volume, rather than providing only final labels.

\subsection{Data Source and Curation}
CT-FlowBench is built on the CT-RATE corpus~\cite{hamamci2024developing} inheriting a vast library of 3D chest CT scans and expert-aligned radiology reports. 
Trajectory synthesis over the full CT-RATE corpus is computationally expensive, so we curate a subset that preserves high reasoning density.
Using heuristic filters, we prioritize cases with higher anatomical diversity, richer diagnostic content, and stronger potential for quantitative assessment.
This strategy concentrates CT-FlowBench on complex, multi-step reasoning trajectories that require cross-slice navigation and precise tool use, while excluding redundant cases that only support single-step reasoning or lack actionable clinical findings.
Details of the curation procedure are provided in Appendix.




\subsection{Task Scenario Definitions.}
We define three complementary functional scenarios spanning fundamental perception to holistic reasoning to evaluate the model’s capabilities in discriminative precision and logical consistency. See the Appendix for details.

\paragraph{Quantitative Analysis.} This scenario assesses the model’s precision in identifying objective physical properties (e.g., dimensions and attenuation values), by discriminating subtle numerical gradients and technical constants from multiple-choice options while filtering out imaging noise.

\paragraph{Spatial Mapping.} This scenario focuses on the recognition of spatial topology and structural adjacency, requiring models to analyze relationships of sequential slices (e.g., contact, encasement, or displacement), and evaluates the ability to translate 3D visual details into precise spatial logic.

\paragraph{Diagnostic Inference.} This scenario requires models to correlate multi-dimensional findings and systemic observations, synthesize localized features into a unified global logic, and select the most coherent clinical inference. This assesses the ability to overcome local information bias through comprehensive holistic reasoning.

\subsection{Trajectory Synthesis}
\label{subsec:trajectory_synthesis}

To bridge the gap between static task definitions and agentic navigation, we employ an Execution-in-the-loop Trajectory Synthesis strategy. Unlike traditional VQA datasets that provide direct mappings from images to labels, CT-FlowBench requires the model to generate a sequence of interleaved thoughts, actions, and observations—collectively termed a ``reasoning trajectory.''

The construction of these trajectories follows an Execution-Feedback Refinement protocol. For each curated case, we utilize a teacher model (e.g., GPT-4o with specialized medical prompting) to explore multiple potential reasoning paths. To ensure the quality and clinical validity of these paths, a trajectory $\mathcal{T}$ is only included in the final benchmark if it satisfies the Procedural Consistency criterion:
\begin{equation}
\forall (a_i, o_i) \in \mathcal{T}, \; \text{val}(o_i \mid \mathcal{V}) \land \text{pred}(\mathcal{T}) = y_{gt}
\end{equation}

In this formulation, $a_i$ and $o_i$ represent the $i$-th action (e.g., a specific slice navigation or ROI crop) and its corresponding observation. The function $\text{is\_valid}$ checks whether the observation $o_i$ is physically grounded and retrievable from the raw volumetric data, while $\text{predict}(\mathcal{T})$ ensures the entire logical chain terminates in the ground-truth diagnosis $y_{gt}$.

In total, we construct CT-FlowBench with 300 evaluation instances and approximately 2000 QA training instances for instruction tuning, where each training instance corresponds to one executable trajectory. The training set consists of two components: (1) trajectory-form instances from CT-Flow (based on a curated subset of CT-RATE) and (2) a subset of the 3D-RAD training data. 
We synthesize trajectories using multiple teacher models (Gemini-3-Pro-Preview/GPT-5.2/Claude-Sonnet-4.5) and apply execution-in-the-loop verification with consistency-based filtering, retaining only instances whose intermediate observations are reproducible via tool execution on the raw volume and whose final answers match the annotations.

%% file: sections/5_experiments.tex
\section{Experiments}
\label{sec:Experiments}
We evaluate CT-Flow on CT-FlowBench and 3D CT VQA dataset, 3D-RAD, to validate its diagnostic accuracy and tool-use autonomy. 
We conduct experiments to compare our agentic framework against state-of-the-art baselines and analyze the efficacy of its tool-mediated reasoning workflow.

\begin{table*}[t]
  \centering
  \caption{Performance comparison on 3D-RAD and CT-FlowBench. For 3D-RAD, the reported metrics represent the average performance across all sub-tasks. For CT-FlowBench, QA, AM, and DD correspond to Quantitative Analysis, Spatial Mapping, and Diagnostic Inference, respectively.}
  \label{tab:main_results}
  \scriptsize
  \setlength{\tabcolsep}{3.5pt} 
  \renewcommand{\arraystretch}{1.1} 
  
  \begin{tabular}{l c cccc c cccc} 
    \toprule
    \multirow{2}{*}{\textbf{Models}} & \multirow{2}{*}{\textbf{Tool-use}}
    & \multicolumn{5}{c}{\textbf{3D-RAD}}
    & \multicolumn{4}{c}{\textbf{CT-FlowBench (Acc. \%)}} \\
    \cmidrule(lr){3-7} \cmidrule(lr){8-11}
    
    & & BLEU-4 & ROUGE-L & B-Score & LLM-Judge & ACC(\%)
    & QA & AM & DD & \textbf{Avg.} \\
    \midrule
    
    \multicolumn{11}{c}{\textcolor{gray}{\textit{Leading-edge Models}}} \\
    
    GPT-5.2 & $\checkmark$ & 22.08 & 26.06 & 85.54 & 18.25 & \underline{63.50} & 35.00 & 40.00 & 37.00 & 37.33 \\
    Gemini-3-Pro-Preview & $\checkmark$ & 29.59 & 35.59 & 89.46 & \textbf{26.38} & 62.59 & \underline{45.00} & 43.00 & \underline{44.00} & 44.00 \\
    Claude-Sonnet-4.5 & $\checkmark$ & 20.14 & 26.81 & 85.71 & 8.85 & 54.83 & 44.00 & 44.00 & 43.00 & 43.67 \\
    Qwen3-VL-235B-A22B-Instruct & $\checkmark$ & 20.55 & 22.62 & 85.44 & 9.52 & 54.21 & 30.00 & 36.00 & 36.00 & 34.00 \\
    GLM4.6-V & $\checkmark$ & 23.23 & 25.47 & 81.59 & 16.71 & 52.04 & 25.00 & 33.00 & 34.00 & 30.67 \\
    
    \midrule
    \multicolumn{11}{c}{\textcolor{gray}{\textit{Specialized Medical VLMs}}} \\
    
    M3D-LaMed-Llama-2-7B & $\times$ & 12.33 & 19.70 & 86.99 & 14.88 & 24.17 & 17.00 & 17.00 & 17.00 & 17.00 \\
    M3D-RAD & $\times$ & 29.76 & \underline{37.39} & \underline{91.30} & \underline{26.00} & 58.00 & 39.00 & 34.00 & 35.00 & 36.00 \\
    Hulu-Med-7B & $\times$ & 12.77 & 23.71 & 86.77 & 22.84 & 61.29 & \textbf{58.00} & \underline{46.00} & 37.00 & \textbf{47.00} \\

    \midrule
    \multicolumn{11}{c}{\textcolor{gray}{\textit{Backbones}}} \\

    Qwen2.5-VL-7B-Instruct & $\checkmark$ & 18.33 & 23.46 & 77.69 & 7.50 & 26.83 & 14.00 & 22.00 & 21.00 & 19.00 \\
    Qwen3-VL-8B-Instruct & $\checkmark$ & 20.89 & 22.31 & 80.07 & 10.77 & 49.06 & 30.00 & 26.00 & 20.00 & 25.33 \\

    \midrule
    \multicolumn{11}{c}{\textcolor{gray}{\textit{Supervised Fine-Tuned Models}}} \\
    \rowcolor{lightblue} \textbf{CT-Flow-7B} & $\checkmark$ & \underline{36.67} & 34.73 & 89.30 & 22.50 & 61.36 & 43.00 & \textbf{52.00} & 38.00 & \underline{44.33} \\
    \rowcolor{lightblue} \textbf{CT-Flow-8B} & $\checkmark$ & \textbf{36.96} & \textbf{37.47} & \textbf{91.65} & 23.63 & \textbf{69.46} & 42.00 & 40.00 & \textbf{47.00} & 43.00 \\
    \bottomrule
  \end{tabular}
\end{table*}

\subsection{Experimental Protocols}
\paragraph{Benchmarks.} The evaluation of CT-Flow is conducted on the proposed CT-FlowBench and a refined subset of 3D-RAD VQA. 
To ensure a balanced and robust assessment, a stratified sampling of 200 instances per task is performed, with option orders randomized for each query to neutralize positional bias (detailed in Appendix).
\paragraph{Baselines.} We benchmark CT-Flow against a diverse set of baselines, including: (i) Leading-edge models (\textit{GPT-5.2}~\cite{openai2025gpt5systemcard}, \textit{Gemini-3-Pro-Preview}~\cite{gemini2023}, \textit{Claude-Sonnet-4.5}, \textit{Qwen3-VL-235B-A22B-Instruct}~\cite{qwen3}, and \textit{GLM4.6-V}~\cite{5team2025glm45agenticreasoningcoding}), (ii) Efficient VLM backbones (Qwen3-VL-8B-Instruct and \textit{Qwen2.5-VL-7B-Instruct})~\cite{qwen2.5}, and (iii) Specialized medical models (\textit{M3D-LaMed-Llama-2-7B}~\cite{bai2024m3d}, \textit{M3D-RAD}~\cite{xin2025med3dvlm}, and \textit{Hulu-Med-7B}~\cite{jiang2025hulu}).
\paragraph{Metrics} We evaluate performance using accuracy on multiple-choice questions. For open-ended tasks, we adopt (i) Automated linguistic metrics, including BLEU-4, ROUGE-L, and BERTScore, to measure semantic and structural alignment; and (ii) LLM-as-a-Judge (average of Deepseek-V3~\cite{deepseekai2025deepseekv32}, Kimi-K2-Thinking, and GPT-OSS-120B~\cite{openai2025gptoss120bgptoss20bmodel}) protocol to provide a high-level assessment.
\paragraph{Implementations.}
All experiments are conducted on 4 NVIDIA H100 GPUs using the LLaMA-Factory~\cite{zheng2024llamafactory} framework. 
For the fine-tuning stage, we perform full-parameter fine-tuning on \textit{Qwen2.5-VL-7B-Instruct} and \textit{Qwen3-VL-8B-Instruct} with a learning rate of $1 \times 10^{-5}$, optimized using DeepSpeed ZeRO-2~\cite{rasley2020deepspeed} with a cosine learning rate decay schedule. 
For model deployment and evaluation, we utilize SGLang~\cite{zheng2024sglangefficientexecutionstructured} to serve the model and conduct inference via its official OpenAI-compatible APIs, ensuring high-throughput and efficient generation.

\begin{figure*}[htbp]
    \centering
    \includegraphics[width=1\linewidth]{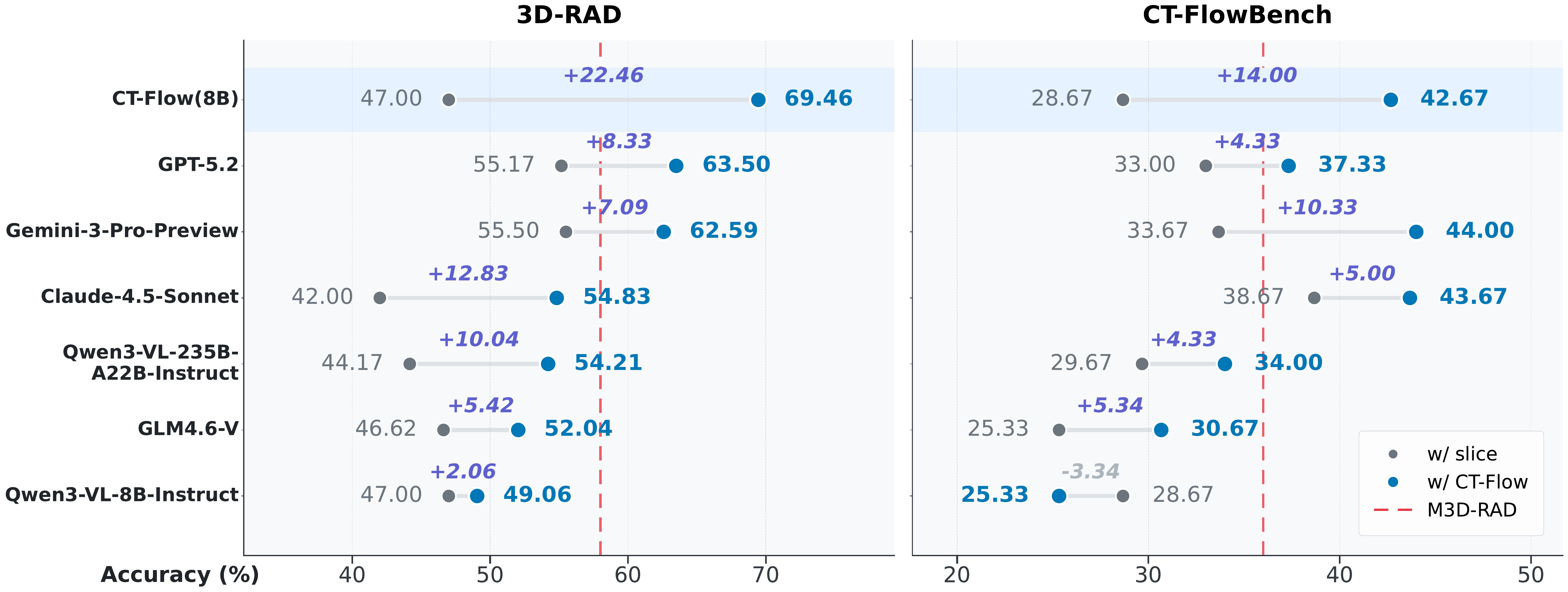}
    \caption{Comparative performance of various models using the CT-Flow framework vs. the slice-based baseline.}
    \label{fig:dumbbell_analysis}
\end{figure*}

\subsection{Comparison with SOTA Methods.}
The experimental results demonstrate the clear superiority of domain-specific supervised fine-tuning (SFT) in elevating diagnostic precision and linguistic alignment. As shown in Table \ref{tab:main_results}, the CT-Flow-8B achieves SOTA performance on the 3D-RAD benchmark with an accuracy of 69.46\%, a significant improvement over its base version and other general-purpose models. This performance gain is mirrored in the linguistic metrics, where SFT variants consistently outperform other models in BLEU-4, ROUGE-L, and BERTScore, indicating that specialized training allows the model to generate reports that are not only factually accurate but also conform to professional clinical standards. The substantial +22.46\% gain visualized in Figure \ref{fig:dumbbell_analysis} further confirms that while general reasoning is a strong foundation, domain-specific adaptation is essential for bridging the gap between general visual understanding and expert-level medical diagnosis.

Simultaneously, our evaluation highlights the transformative impact of the CT-Flow agentic framework on general-purpose frontier models, while exposing the increased logical complexity of CT-FlowBench. Through tool-mediated reasoning, general models such as GPT-5.2 and Gemini-3-Pro-Preview effectively surpass specialized medical models like M3D-RAD (58.00\%), achieving accuracies of 63.50\% and 62.59\% on 3D-RAD, respectively. However, the performance on CT-FlowBench is notably lower across all models, with the highest average accuracy reaching only 44.33\%. This discrepancy underlines the benchmark's challenge in evaluating multi-stage tool-use rather than simple visual recognition. Smaller models like Qwen3-8B even experience a slight performance regression on this benchmark, suggesting that the cognitive load of autonomous tool-calling and differential diagnosis requires either high-scale reasoning capabilities or targeted fine-tuning to maintain logical consistency throughout the workflow.

\subsection{Analysis of Tool-mediated Reasoning.}
The experimental results illustrated in Figure~\ref{fig:dumbbell_analysis} demonstrate a universal performance elevation across the 3D-RAD benchmark, confirming the efficacy of the CT-Flow framework in significantly enhancing diagnostic accuracy. Most notably, the integration of our agentic tool-use workflow enables the CT-Flow (SFT) model to achieve a state-of-the-art accuracy of 69.46\%, which constitutes a substantial +22.46\% improvement over the slice-based baseline. This trend extends to general-purpose frontier models such as GPT-5.2 and Claude-Opus, which exhibit gains of +8.33\% and +12.83\% respectively. By leveraging tool-mediated reasoning, these general-domain models are able to significantly outperform specialized medical models like M3D-RAD (58.00\%), proving that an agentic approach can effectively bridge the domain expertise gap in medical imaging without the need for exhaustive medical-specific pre-training.

However, the results on the CT-FlowBench reveal that the benefits of tool-mediated reasoning are closely tied to model scale and the inherent complexity of the reasoning task. While high-capacity models like Gemini-3-Pro and GPT-5.2 continue to show strong positive gains (+10.33\% and +14.00\% respectively), the smaller Qwen3-8B model exhibits a performance regression of -3.34\%. This divergence suggests that the multi-step diagnostic logic required by the benchmark—encompassing Spatial Mapping, Quantitative Analysis, and Diagnostic Inference—places a high cognitive demand on instruction-following and strategic planning that may exceed the zero-shot capabilities of smaller backbones. Despite this scaling effect, the consistent upward trajectory across the majority of models confirms that decomposing complex queries into tool-mediated steps is a far more robust strategy for 3D medical analysis than traditional static slice feeding, provided the underlying model possesses sufficient reasoning depth to manage the autonomous tool-calling workflow.

\subsection{Tool-use Performance \& Ablations}

\begin{table*}[htbp]
  \centering
  \small
  \setlength{\tabcolsep}{4pt} 
  \renewcommand{\arraystretch}{1.1}
  
    \begin{tabular}{lcccccc}
      \toprule
      \multirow{2}{*}{\textbf{Models}} &
      \multicolumn{3}{c}{\textbf{3D-RAD}} &
      \multicolumn{3}{c}{\textbf{CT-FlowBench}} \\
      \cmidrule(lr){2-4} \cmidrule(lr){5-7}
      & Calls & Name Errors & Args Errors
      & Calls & Name Errors & Args Errors \\
      \midrule

      GPT-5.2 & 4.133 & 0.006 & 0.056 & 7.193 & 0.003 & 0.108 \\
      Gemini-3-Pro-Preview & 5.088 & 0.008 & 0.156 & 6.667 & 0.100 & 0.451 \\
      Claude-Sonnet-4.5 & 5.930 & 0.002 & 0.092 & 9.490 & 0.017 & 0.407 \\
      Qwen3-VL-235B-A22B-Instruct & 6.072 & 0.085 & 0.190 & 6.719 & 0.224 & 0.453 \\
      Qwen3-VL-8B-Instruct & 5.963 & 0.782 & 0.211 & 11.196 & 0.969 & 0.385 \\
      GLM4.6-V & 4.030 & 0.027 & 0.078 & 6.314 & 0.093 & 0.204 \\
      
      \midrule
      
      \textbf{CT-Flow-7B} & 4.013 & 0.007 & 0.018 & 6.170 & 0.007 & 0.033 \\
      \textbf{CT-Flow-8B} & 4.248 & 0.007 & 0.057 & 7.480 & 0.027 & 0.282 \\

      \bottomrule
    \end{tabular}

  \caption{Tool usage and error statistics on 3D-RAD and CT-FlowBench. \textit{Calls}, \textit{Name Errors}, and \textit{Args Errors} denote the average number of tool calls, tool-name errors and tool-argument errors per case, respectively.}
  \label{tab:tool_usage_hallucination}
\end{table*}

\paragraph{Tool-use Performance Analysis.}
Table \ref{tab:tool_usage_hallucination} provides a comprehensive statistical breakdown of tool-calling behaviors across various models on the 3D-RAD and CT-FlowBench datasets. The results indicate a direct correlation between task complexity and the frequency of tool interactions. For instance, Claude-Sonnet-4.5 and Qwen3-VL-235B exhibit a high average number of tool calls, particularly in the CT-FlowBench, where complex diagnostic workflows require multi-step reasoning and iterative data retrieval.

Regarding reliability, GPT-5.2 sets the benchmark for instruction following, maintaining near-zero tool-name errors and minimal argument hallucinations. This stability is crucial in clinical settings where precise parameter input is non-negotiable. Conversely, smaller-scale models such as Qwen3-VL-8B-Instruct show significant degradation in performance; the high frequency of name errors (0.782) suggests that these models struggle to maintain the logical consistency required for long-chain tool manipulation in 3D medical spaces. These findings underscore that while vision-language alignment is necessary, the capability to orchestrate professional tools is distinctly scaling property of larger, more sophisticated.

\paragraph{Ablation Study on Tool Categories.}
To evaluate the structural integrity and necessity of our proposed toolset, we conducted a systematic ablation study by categorizing the tools into four functional tiers. In our framework, Data Ingestion serves as the foundational prerequisite for any task execution, while the other three categories are incrementally added or removed to observe success rates.

\begin{figure}
    \centering
    \includegraphics[width=1\linewidth]{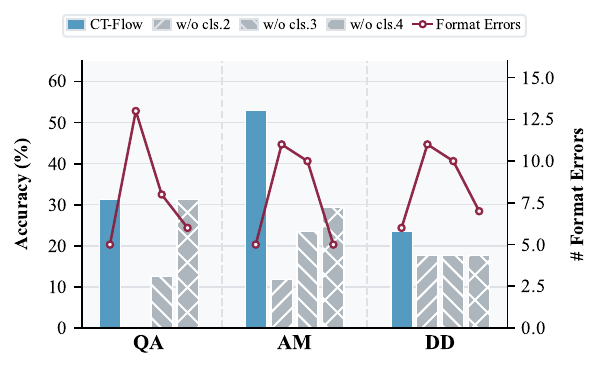}
    \caption{Performance impact of tool category ablation. Bars indicate accuracy (\%) and the red line tracks format errors. Removing specific tool classes (cls. 2-4) leads to decreased diagnostic accuracy and increased errors across all tasks, validating the necessity of the full hierarchical toolset.}
    \label{fig:ablation}
\end{figure}

The experimental results~\ref{fig:ablation} demonstrate that each category plays an indispensable role in the clinical reasoning pipeline. The removal of Advanced Analysis tools leads to a failure in synthesizing quantitative clinical metrics, such as calculating flow velocities or volumetric ratios, which are essential for definitive diagnosis. Without Detailed Observation capabilities, the system's sensitivity to micro-lesions and subtle anatomical anomalies drops sharply, proving that global context alone is insufficient for high-precision radiology. Furthermore, the absence of Global Navigation results in ``spatial disorientation'' within the model’s reasoning process, as it loses the ability to efficiently index and transition between disparate 3D slices or sequences.
Ultimately, the ablation results confirm that our toolset is not merely a collection of independent functions but a coherent, hierarchical system. We conclude that all four tool categories are essential and rationally designed for addressing the inherent challenges of 3D medical images.

%% file: sections/6_conclusion.tex
\section{Conclusion}
\label{sec:Conclusion}

We present CT-Flow, a novel agentic framework that shifts the paradigm of 3D CT analysis from passive visual ingestion to protocol-driven Active Probing. 
By integrating the MCP and the trajectory-based CT-FlowBench, our approach enables an orchestrator to dynamically invoke specialized clinical tools, effectively overcoming the information bottlenecks of traditional 3D-LVLMs. 
Empirical results demonstrate that CT-Flow not only achieves state-of-the-art performance across diagnostic benchmarks but also provides a transparent and traceable reasoning process that aligns with real-world radiological workflows, offering a promising path toward more interactive and reliable AI-assisted medicine.

%% file: sections/7_limitation.tex
\section*{Limitations}
\label{sec:Related Work}

Despite the significant advancements made by CT-Flow in agentic clinical interpretation, this work has certain limitations. First, the current model relies primarily on supervised fine-tuning (SFT) on CT-FlowBench and has not yet integrated reinforcement learning (RL) to further optimize decision trajectories or minimize redundant tool invocations. Recognizing the potential of Reinforcement Learning from Clinical Feedback (RLCF) in achieving expert-level diagnostic efficiency, we plan to incorporate algorithms such as Proximal Policy Optimization (PPO) or Direct Preference Optimization (DPO) in future iterations to refine reasoning chains. Second, while this iterative multi-step paradigm provides in-depth analysis, its inference latency is inherently higher than that of traditional single-pass models, which may pose challenges in time-sensitive clinical scenarios such as acute stroke triage. Consequently, improving inference efficiency through parallel processing or trajectory pruning, while maintaining diagnostic precision and transparency, will be a key focus of our future research.

%% file: sections/8_ethical.tex
\section*{Ethical Considerations}

\paragraph{Clinical Oversight and Agency}
While CT-Flow transitions from passive inference to active, tool-mediated orchestration, it is designed strictly as a clinical decision support system (CDSS) rather than an autonomous diagnostic entity. The ``Thought-Action Trajectories'' generated by the model are intended to provide clinicians with a transparent reasoning path and tool-verified evidence. However, the risk of ``action hallucinations'' in which the agent might invoke incorrect tools or misinterpret tool outputs persists. Therefore, all automated tool sequences and final diagnostic suggestions must be reviewed and validated by certified radiologists. We advocate for a ``human-in-the-loop'' deployment to ensure patient safety.

\paragraph{Transparency and Accountability}
By leveraging the Model Context Protocol (MCP), CT-Flow provides an auditable trail of how a diagnostic conclusion was reached. This traceability is a key ethical safeguard, allowing clinicians to inspect which slices were visualized and which radiomic features were extracted. Despite this, users should remain vigilant regarding the model's performance across different patient demographics and imaging protocols.

\paragraph{Data Privacy and Provenance}
CT-FlowBench is constructed using de-identified, publicly available 3D CT datasets. Throughout the curation and instruction-tuning process, we have strictly adhered to ethical guidelines to ensure that no Protected Health Information (PHI) is included. All data processing workflows are compliant with standard medical data privacy regulations (e.g., HIPAA), and the framework is designed to be deployed in secure, local clinical environments to prevent unauthorized data exposure.

%% file: sections/10_appendix.tex
\newpage
\section{Appendix}

\subsection{Provided Tools in CT-Flow}
To facilitate a streamlined and rigorous medical image analysis pipeline, CT-Flow provides a comprehensive suite of specialized tools categorized into four functional modules. As summarized in Table \ref{tab:medical_tools}, these modules are designed to mirror the natural clinical workflow—moving from raw data validation to holistic screening, followed by precise local quantification, and finally culminating in AI-driven prognostic modeling. This modular architecture ensures both the reproducibility of the analysis and the scalability of the system for large-scale clinical studies.
\begin{table*}[htbp]
\centering
\caption{Classification and Functional Description of Medical Imaging Analysis Tools}
\label{tab:medical_tools}
\small 
\begin{tabularx}{\textwidth}{l p{4.5cm} X}
\toprule
\textbf{Category} & \textbf{Included Tools} & \textbf{Clinical Objective \& Description} \\
\midrule
\textbf{I. Data Ingestion} & 
\texttt{load\_data}, \texttt{load\_mask}, \texttt{inspect\_metadata}, \texttt{inspect\_mask\_labels}, \texttt{search\_anatomy\_names}, \texttt{list\_window\_presets} & 
\textbf{Data Integrity:} Establishes the workspace and validates metadata (modality, spacing, labels) to ensure subsequent analysis is performed on accurate objects. \\
\addlinespace[0.8em]

\textbf{II. Global Observation} & 
\texttt{view\_montage}, \texttt{view\_mip}, \texttt{view\_minip}, \texttt{view\_avgip} & 
\textbf{Macro-Navigation:} Provides a ``bird's-eye view'' using projection (MIP/MinIP) or array layouts to help users rapidly screen for high-density (nodules) or low-density (airways) abnormalities. \\
\addlinespace[0.8em]

\textbf{III. Detailed Measurement} & 
\texttt{view\_slice}, \texttt{view\_ortho}, \texttt{measure\_distance}, \texttt{measure\_max\_diameter}, \texttt{find\_organ\_center}, \texttt{extract\_vessel\_centerline}, \texttt{auto\_crop\_body}, \texttt{edit\_geometry} & 
\textbf{Spatial Localization:} The most high-frequency module. Combines slice-by-slice exploration with 3D multi-planar reconstruction (MPR) and precise physical quantification of lesion size and geometry. \\
\addlinespace[0.8em]

\textbf{IV. Advanced Analytics} & 
\texttt{segment\_total\_anatomy}, \texttt{analyze\_hu\_distribution}, \texttt{analyze\_shape\_properties}, \texttt{extract\_radiomics\_signature}, \texttt{visualize\_radiomics\_chart}, \texttt{analyze\_lesion\_texture} & 
\textbf{Quantitative Biomarkers:} Post-processing stage utilizing AI and statistics to transform visual data into structured radiomic features for prognostic modeling and malignancy assessment. \\
\bottomrule
\end{tabularx}
\end{table*}

\subsection{Details of Dataset Construction}

This appendix provides a comprehensive description of the systematic pipeline developed to transform unstructured clinical CT reports into a high-precision, tool-calling benchmark. The construction logic is designed to move beyond simple question-answering, focusing instead on the agent's ability to execute complex, multi-step clinical reasoning within a constrained action space.

\subsubsection{Multi-Agent Annotation and Taxonomic Categorization}
The process begins with the raw ingestion of clinical reports, which are processed by a specialized \textit{Medical Data Annotation Expert} agent. The primary objective of this stage is to filter and categorize reports based on their potential to challenge an AI agent’s reasoning capabilities through a weighted scoring system.

The agent evaluates reports based on three primary features: \textbf{Numerical Presence} (essential for quantitative precision tasks), \textbf{Anatomical Diversity} (requiring cross-regional spatial logic), and \textbf{Clinical Uncertainty} (necessitating autonomous decision-making). Based on these scores, reports are classified into three types: \textbf{Type A (Atomic)} for simple fact retrieval, \textbf{Type B (Logical)} for spatial relational tasks, and \textbf{Type C (Autonomous)} for complex reasoning. Reports containing explicit diagnostic uncertainty or follow-up recommendations are prioritized as Type C to ensure the benchmark includes cases that simulate a radiologist’s diagnostic synthesis.

\subsubsection{High-Precision Ground Truth Synthesis}
Once a report is categorized, a \textit{Senior Radiologist Agent} acts as a data architect to convert clinical findings into a structured JSON ``Gold Standard''. This transformation is governed by the \textbf{Hard Evidence Principle}, which dictates that the agent must only extract findings supported by explicit measurements, specific anatomical descriptors, or clear Hounsfield Unit (HU) density values.

Furthermore, this stage involves \textbf{Tool-Logic Mapping}. For every ground-truth fact, the radiologist agent defines the specific sequence of tools from the internal library required for verification. For instance, a finding regarding a pulmonary nodule is programmatically linked to \texttt{segment\_total\_anatomy} and \texttt{measure\_max\_diameter} functions. This ensures that the benchmark evaluates not only the final answer but also the validity of the agent's execution path.

\subsubsection{Clinical Scenario Domain Configuration}
To ensure a broad evaluation across radiologic specialties, each benchmark task is assigned to a specific clinical scenario. This allows for granular analysis of where an AI agent may excel or fail.
\begin{itemize}
    \item \textbf{Scenario A (Quantitative Precision):} Focuses on RECIST 1.1 compliance, absolute metric accuracy, and HU density analysis to test mathematical and volumetric precision.
    \item \textbf{Scenario B (Surgical Mapping):} Focuses on the assessment of vascular encasement angles and tumor-vessel cleavage planes, testing the agent's ability to provide preoperative spatial intelligence.
    \item \textbf{Scenario C (Diagnostic Synthesis):} Focuses on systemic staging (TNM), enhancement pattern analysis, and complex pattern recognition across multiple organ systems.
\end{itemize}

\subsubsection{The ``Blind Verification'' Instruction Strategy}
A critical challenge in evaluating AI agents is preventing ``textual cheating'', where the agent uses its internal linguistic knowledge to guess findings from the report text. To mitigate this, we implemented a \textbf{Blind Verification} strategy. 

During instruction generation, the agent is forbidden from using explicit pathology labels, such as ``the pancreatic tumor.'' Instead, it must describe the target using \textbf{Spatial Landmark Referencing} (e.g., ``the lesion 10mm inferior to the Splenic Vein''). By removing explicit labels, the agent is forced to use visualization and segmentation tools to navigate the 3D volume and locate the target before it can perform any analysis. This ensures the benchmark measures true tool-calling and image-grounding capabilities.

\subsubsection{Action Space Constraints and SOP Validation}
The final stage of construction ensures that every generated task is solvable within the \textbf{Action Space} of the predefined CT-AI Tool Library. For every task, the pipeline generates a \textbf{Standard Operating Procedure (SOP)} ground truth, representing the optimal sequence of tool calls a human expert would utilize to reach the conclusion.

This SOP serves as a trajectory gold standard. By comparing the AI agent’s actual tool-calling sequence against this expert-defined SOP, we can evaluate the efficiency, medical logic, and safety of the agent’s diagnostic process, rather than relying solely on the accuracy of the final multiple-choice selection.

\subsection{Comparative Analysis: 2D Slice-based vs. CT-Flow Tool Processing}

The experimental results in Table~\ref{tab:performance_diff} and Table~\ref{tab:scenario_performance} highlight a critical performance gap between standard 2D slice-based vision processing and our proposed \textit{CT-Flow} tool-enhanced framework. While baseline models process individual 2D slices without a global spatial understanding, the \textit{CT-Flow} tool enables these models to integrate 3D spatial-contextual information.

\subsubsection{Bridging the 2D-to-3D Gap (Table 1)}
Analysis of the 3D Rad 1.2k dataset reveals that the lack of spatial context in the slice-based baseline severely limits diagnostic and descriptive performance:

\begin{itemize}
    \item \textbf{Overcoming Information Fragmentation:} Baseline models (Gemini-3-Pro, GPT-5.2, etc.) operating on individual slices struggle with Tasks 1--3 (Report Generation). For example, \textbf{Gemini-3-Pro}'s BLEU and Rouge scores more than double when utilizing the \textit{CT-Flow} tool (BLEU: $11.72 \rightarrow 26.85$; Rouge: $15.81 \rightarrow 38.29$). This suggests that while individual slices contain local features, the \textit{tool} provides the necessary spatial continuity to generate coherent clinical reports.
    \item \textbf{Diagnostic Accuracy through Spatial Synthesis:} In Tasks 4--6, which require identifying volumetric features, the slice-based approach shows significant limitations. The integration of \textit{CT-Flow} allows models to synthesize information across slices, leading to substantial accuracy gains. For \textbf{Claude-Opus}, the tool-enhanced version improves Task 5 accuracy by 23.19\%, demonstrating that spatial relationships are essential for accurate medical classification.
    \item \textbf{Efficiency vs. Specialization:} Notably, the tool-enhanced general-purpose models significantly outperform \textbf{M3D}, a model natively designed for 3D volumes. This indicates that the combination of high-level reasoning in LLMs and our spatial-contextual \textit{tool} is more effective than specialized architectures that process 3D data in a black-box manner.
\end{itemize}

\subsubsection{Scenario-Based Robustness (Table 2)}
Table~\ref{tab:scenario_performance} further illustrates how the tool-based approach enhances model stability across different evaluation scenarios compared to the slice-only method:

\begin{itemize}
    \item \textbf{Consistent Enhancement:} Across all scenarios, the transition from 2D slice processing to \textit{CT-Flow} tool integration leads to an upward trend. \textbf{Gemini-3-Pro} achieved an average improvement of 10.33 points, reaching a peak of 45.00 in Scenario 1. This confirms that the tool provides a robust spatial prior that remains effective regardless of the specific scenario configuration.
    \item \textbf{Addressing Slice-Level Ambiguity:} The baseline models often exhibit high variance between scenarios (e.g., \textbf{Qwen3-VL-8B} dropping from 31.00 to 22.00). The \textit{CT-Flow} tool acts as a ``spatial stabilizer'', helping models resolve ambiguities that occur when a lesion or anatomical structure is only partially visible or appears differently across various 2D slices.
    \item \textbf{Fine-tuning vs. Tool Integration:} Interestingly, the \textit{Fine-tuned} model—which was explicitly trained to handle these volumes—shows the highest Scenario 3 performance (46.00). However, the fact that \textbf{Gemini-3-Pro + CT-Flow} (44.00) reaches a similar level without scenario-specific training underscores the power of providing the model with a contextual tool rather than just raw image slices.
\end{itemize}

\subsubsection{Summary of Findings}
The data provides strong empirical evidence that 2D slice-based processing is insufficient for complex 3D medical tasks. By utilizing the \textit{CT-Flow} tool, models can leverage spatial-contextual information to achieve a more holistic understanding of the 1.2k dataset, resulting in performance that rivals or exceeds natively 3D-aware models and approaches the ceiling of fine-tuned systems.

\begin{table*}[!h]
  \centering
  \caption{Performance comparison on the 3D Rad 1.2k dataset. M3D natively supports 3D volumes, whereas other baseline models process individual 2D slices. Models denoted with \textit{w CT Lens} leverage our proposed CTLens framework to integrate spatial-contextual information.}
  \label{tab:performance_diff}
  \footnotesize 
  \setlength{\tabcolsep}{2.5pt}
  \renewcommand{\arraystretch}{1.1}
  \begin{tabular}{l ccc ccc ccc ccc c}
    \toprule
    \multirow{2.5}{*}{Model} & \multicolumn{3}{c}{Task 1} & \multicolumn{3}{c}{Task 2} & \multicolumn{3}{c}{Task 3} & Task 4 & Task 5 & Task 6 & \multirow{2.5}{*}{\textbf{Avg.}} \\
    \cmidrule(lr){2-4} \cmidrule(lr){5-7} \cmidrule(lr){8-10} \cmidrule(lr){11-11} \cmidrule(lr){12-12} \cmidrule(lr){13-13}
    & BLEU & Rouge & Bert & BLEU & Rouge & Bert & BLEU & Rouge & Bert & acc & acc & acc & \\
    \midrule
    Gemini-3-Pro & 11.72 & 15.81 & 73.30 & 12.47 & 19.36 & 75.35 & 12.05 & 12.04 & 53.30 & 61.00 & 42.50 & 63.00 & 55.50 \\
    \quad \textit{w CT-Flow} & \textbf{26.85} & \textbf{38.29} & 87.94 & 25.95 & 34.38 & 87.67 & \textbf{35.98} & 34.10 & 92.78 & 60.50 & 52.26 & 75.00 & 62.59 \\
    & {\scriptsize (\textcolor{upgreen}{$\uparrow$15.13})} & {\scriptsize (\textcolor{upgreen}{$\uparrow$22.48})} & {\scriptsize (\textcolor{upgreen}{$\uparrow$14.64})} & {\scriptsize (\textcolor{upgreen}{$\uparrow$13.48})} & {\scriptsize (\textcolor{upgreen}{$\uparrow$15.02})} & {\scriptsize (\textcolor{upgreen}{$\uparrow$12.32})} & {\scriptsize (\textcolor{upgreen}{$\uparrow$23.93})} & {\scriptsize (\textcolor{upgreen}{$\uparrow$22.06})} & {\scriptsize (\textcolor{upgreen}{$\uparrow$39.48})} & {\scriptsize ($\downarrow$0.50)} & {\scriptsize (\textcolor{upgreen}{$\uparrow$9.76})} & {\scriptsize (\textcolor{upgreen}{$\uparrow$12.00})} & {\scriptsize (\textcolor{upgreen}{$\uparrow$7.09})} \\
    \addlinespace[4pt]
    
    Gpt-5.2 & 9.51 & 16.67 & 83.77 & 10.72 & 19.52 & 84.34 & 1.08 & 5.03 & 81.37 & 59.50 & 36.00 & 70.00 & 55.17 \\
    \quad \textit{w CT-Flow} & 19.68 & 25.38 & 84.60 & 20.42 & 26.61 & 85.08 & 26.14 & 26.19 & 86.94 & 70.00 & 43.00 & 77.50 & 63.50 \\
    & {\scriptsize (\textcolor{upgreen}{$\uparrow$10.17})} & {\scriptsize (\textcolor{upgreen}{$\uparrow$8.71})} & {\scriptsize (\textcolor{upgreen}{$\uparrow$0.83})} & {\scriptsize (\textcolor{upgreen}{$\uparrow$9.70})} & {\scriptsize (\textcolor{upgreen}{$\uparrow$7.09})} & {\scriptsize (\textcolor{upgreen}{$\uparrow$0.74})} & {\scriptsize (\textcolor{upgreen}{$\uparrow$25.06})} & {\scriptsize (\textcolor{upgreen}{$\uparrow$21.16})} & {\scriptsize (\textcolor{upgreen}{$\uparrow$5.57})} & {\scriptsize (\textcolor{upgreen}{$\uparrow$10.50})} & {\scriptsize (\textcolor{upgreen}{$\uparrow$7.00})} & {\scriptsize (\textcolor{upgreen}{$\uparrow$7.50})} & {\scriptsize (\textcolor{upgreen}{$\uparrow$8.33})} \\
    \addlinespace[4pt]

    Claude-Opus & 5.19 & 14.85 & 82.43 & 5.72 & 16.03 & 83.62 & 6.65 & 19.67 & 83.01 & 54.00 & 14.50 & 57.50 & 42.00 \\
    \quad \textit{w CT-Flow} & 16.25 & 25.38 & 84.92 & 16.73 & 24.39 & 84.49 & 27.45 & 30.67 & 87.72 & 60.80 & 37.69 & 65.99 & 54.83 \\
    & {\scriptsize (\textcolor{upgreen}{$\uparrow$11.06})} & {\scriptsize (\textcolor{upgreen}{$\uparrow$10.53})} & {\scriptsize (\textcolor{upgreen}{$\uparrow$2.49})} & {\scriptsize (\textcolor{upgreen}{$\uparrow$11.01})} & {\scriptsize (\textcolor{upgreen}{$\uparrow$8.36})} & {\scriptsize (\textcolor{upgreen}{$\uparrow$0.87})} & {\scriptsize (\textcolor{upgreen}{$\uparrow$20.80})} & {\scriptsize (\textcolor{upgreen}{$\uparrow$11.00})} & {\scriptsize (\textcolor{upgreen}{$\uparrow$4.71})} & {\scriptsize (\textcolor{upgreen}{$\uparrow$6.80})} & {\scriptsize (\textcolor{upgreen}{$\uparrow$23.19})} & {\scriptsize (\textcolor{upgreen}{$\uparrow$8.49})} & {\scriptsize (\textcolor{upgreen}{$\uparrow$12.83})} \\
    \addlinespace[4pt]

    Qwen3-VL-235B & 21.15 & 25.88 & 85.51 & 15.58 & 19.21 & 85.24 & 3.60 & 6.39 & 84.47 & 55.50 & 26.00 & 51.00 & 44.17 \\
    \quad \textit{w CT-Flow} & 22.43 & 24.68 & 84.56 & 21.48 & 22.98 & 84.18 & 17.75 & 20.19 & 87.58 & 58.33 & 38.50 & 65.79 & 54.21 \\
    & {\scriptsize (\textcolor{upgreen}{$\uparrow$1.28})} & {\scriptsize ($\downarrow$1.20)} & {\scriptsize ($\downarrow$0.95)} & {\scriptsize (\textcolor{upgreen}{$\uparrow$5.90})} & {\scriptsize (\textcolor{upgreen}{$\uparrow$3.77})} & {\scriptsize ($\downarrow$1.06)} & {\scriptsize (\textcolor{upgreen}{$\uparrow$14.15})} & {\scriptsize (\textcolor{upgreen}{$\uparrow$13.80})} & {\scriptsize (\textcolor{upgreen}{$\uparrow$3.11})} & {\scriptsize (\textcolor{upgreen}{$\uparrow$2.83})} & {\scriptsize (\textcolor{upgreen}{$\uparrow$12.50})} & {\scriptsize (\textcolor{upgreen}{$\uparrow$14.79})} & {\scriptsize (\textcolor{upgreen}{$\uparrow$10.04})} \\
    \addlinespace[4pt]

    GLM4.6-V & 18.26 & 26.64 & 86.29 & 19.00 & 25.41 & 86.19 & 9.54 & 14.81 & 89.13 & 63.64 & 13.07 & 63.16 & 46.62 \\
    \quad \textit{w CT-Flow} & 23.22 & 26.98 & 81.27 & 22.90 & 26.17 & 79.57 & 23.58 & 23.26 & 83.94 & 55.50 & 37.84 & 62.77 & 52.04 \\
    & {\scriptsize (\textcolor{upgreen}{$\uparrow$4.96})} & {\scriptsize (\textcolor{upgreen}{$\uparrow$0.34})} & {\scriptsize ($\downarrow$5.02)} & {\scriptsize (\textcolor{upgreen}{$\uparrow$3.90})} & {\scriptsize (\textcolor{upgreen}{$\uparrow$0.76})} & {\scriptsize ($\downarrow$6.62)} & {\scriptsize (\textcolor{upgreen}{$\uparrow$14.04})} & {\scriptsize (\textcolor{upgreen}{$\uparrow$8.45})} & {\scriptsize ($\downarrow$5.19)} & {\scriptsize ($\downarrow$8.14)} & {\scriptsize (\textcolor{upgreen}{$\uparrow$24.77})} & {\scriptsize ($\downarrow$0.39)} & {\scriptsize (\textcolor{upgreen}{$\uparrow$5.41})} \\
    \addlinespace[4pt]

    Qwen3-VL-8B & 16.48 & 20.09 & 80.52 & 15.48 & 18.71 & 80.50 & 2.13 & 4.70 & 82.09 & 58.00 & 25.50 & 57.50 & 47.00 \\
    \quad \textit{w CT-Flow} & 18.97 & 23.98 & 78.53 & 17.97 & 22.01 & 72.17 & 25.74 & 20.95 & 89.50 & 39.13 & 42.39 & 65.66 & 49.06 \\
    & {\scriptsize (\textcolor{upgreen}{$\uparrow$2.49})} & {\scriptsize (\textcolor{upgreen}{$\uparrow$3.89})} & {\scriptsize ($\downarrow$1.99)} & {\scriptsize (\textcolor{upgreen}{$\uparrow$2.49})} & {\scriptsize (\textcolor{upgreen}{$\uparrow$3.30})} & {\scriptsize ($\downarrow$8.33)} & {\scriptsize (\textcolor{upgreen}{$\uparrow$23.61})} & {\scriptsize (\textcolor{upgreen}{$\uparrow$16.25})} & {\scriptsize (\textcolor{upgreen}{$\uparrow$7.41})} & {\scriptsize ($\downarrow$18.87)} & {\scriptsize (\textcolor{upgreen}{$\uparrow$16.89})} & {\scriptsize (\textcolor{upgreen}{$\uparrow$8.16})} & {\scriptsize (\textcolor{upgreen}{$\uparrow$2.06})} \\
    \addlinespace[4pt]

    fine-tuned & 26.84 & 36.57 & 88.22 & 23.42 & 34.50 & 86.48 & 20.81 & 24.28 & 59.89 & \textbf{75.40} & \textbf{54.00} & \textbf{78.99} & \textbf{69.46} \\
    & {\scriptsize (\textcolor{upgreen}{$\uparrow$10.36})} & {\scriptsize (\textcolor{upgreen}{$\uparrow$16.48})} & {\scriptsize (\textcolor{upgreen}{$\uparrow$7.70})} & {\scriptsize (\textcolor{upgreen}{$\uparrow$7.94})} & {\scriptsize (\textcolor{upgreen}{$\uparrow$15.79})} & {\scriptsize (\textcolor{upgreen}{$\uparrow$5.98})} & {\scriptsize (\textcolor{upgreen}{$\uparrow$18.68})} & {\scriptsize (\textcolor{upgreen}{$\uparrow$19.58})} & {\scriptsize ($\downarrow$22.20)} & {\scriptsize (\textcolor{upgreen}{$\uparrow$17.40})} & {\scriptsize (\textcolor{upgreen}{$\uparrow$28.50})} & {\scriptsize (\textcolor{upgreen}{$\uparrow$21.49})} & {\scriptsize (\textcolor{upgreen}{$\uparrow$22.46})} \\
    \midrule
    M3D & 12.15 & 19.04 & 84.98 & 11.80 & 20.68 & 85.46 & 13.05 & 19.39 & 90.54 & 22.50 & 27.00 & 23.00 & 24.17 \\
    3D-RAD & 26.33 & 34.83 & \textbf{89.41} & \textbf{31.66} & \textbf{39.89} & \textbf{89.90} & 31.29 & \textbf{37.46} & \textbf{94.58} & 72.50 & 36.00 & 65.50 & 58.00 \\
    \bottomrule
  \end{tabular}
\end{table*}

\begin{table}[!h]
  \centering
  \caption{Performance comparison across different scenarios. Models denoted with \textit{w CT-Flow} show the integration of spatial-contextual information.}
  \label{tab:scenario_performance}
  
  \definecolor{upgreen}{rgb}{0.0, 0.5, 0.0}
  
  \resizebox{\columnwidth}{!}{
    \footnotesize
    \setlength{\tabcolsep}{4pt}
    \renewcommand{\arraystretch}{1.3}
    \begin{tabular}{l ccc c}
      \toprule
      Model & Scenario 1 & Scenario 2 & Scenario 3 & \textbf{Ave.} \\
      \midrule
      Gemini-3-Pro & 35.00 & 35.00 & 31.00 & 33.67 \\
      \quad \textit{w CT-Flow} & \textbf{45.00} & 43.00 & 44.00 & \textbf{44.00} \\
      & {\scriptsize (\textcolor{upgreen}{$\uparrow$10.00})} & {\scriptsize (\textcolor{upgreen}{$\uparrow$8.00})} & {\scriptsize (\textcolor{upgreen}{$\uparrow$13.00})} & {\scriptsize (\textcolor{upgreen}{$\uparrow$10.33})} \\
      \addlinespace[4pt]
      
      Gpt-5.2 & 26.00 & 36.00 & 37.00 & 33.00 \\
      \quad \textit{w CT-Flow} & 35.00 & 40.00 & 37.00 & 37.33 \\
      & {\scriptsize (\textcolor{upgreen}{$\uparrow$9.00})} & {\scriptsize (\textcolor{upgreen}{$\uparrow$4.00})} & {\scriptsize (0.00)} & {\scriptsize (\textcolor{upgreen}{$\uparrow$4.33})} \\
      \addlinespace[4pt]

      Claude & 43.00 & 39.00 & 34.00 & 38.67 \\
      \quad \textit{w CT-Flow} & 44.00 & \textbf{44.00} & 43.00 & 43.67 \\
      & {\scriptsize (\textcolor{upgreen}{$\uparrow$1.00})} & {\scriptsize (\textcolor{upgreen}{$\uparrow$5.00})} & {\scriptsize (\textcolor{upgreen}{$\uparrow$9.00})} & {\scriptsize (\textcolor{upgreen}{$\uparrow$5.00})} \\
      \addlinespace[4pt]

      Qwen3-VL-235B & 33.00 & 30.00 & 26.00 & 29.67 \\
      \quad \textit{w CT-Flow} & 30.00 & 36.00 & 36.00 & 34.00 \\
      & {\scriptsize ($\downarrow$3.00)} & {\scriptsize (\textcolor{upgreen}{$\uparrow$6.00})} & {\scriptsize (\textcolor{upgreen}{$\uparrow$10.00})} & {\scriptsize (\textcolor{upgreen}{$\uparrow$4.33})} \\
      \addlinespace[4pt]

      GLM4.6-V & 25.00 & 24.00 & 27.00 & 25.33 \\
      \quad \textit{w CT-Flow} & 25.00 & 33.00 & 34.00 & 30.67 \\
      & {\scriptsize (0.00)} & {\scriptsize (\textcolor{upgreen}{$\uparrow$9.00})} & {\scriptsize (\textcolor{upgreen}{$\uparrow$7.00})} & {\scriptsize (\textcolor{upgreen}{$\uparrow$5.33})} \\
      \addlinespace[4pt]

      Qwen3-VL-8B & 31.00 & 33.00 & 22.00 & 28.67 \\
      \quad \textit{w CT-Flow} & 30.00 & 26.00 & 20.00 & 25.33 \\
      & {\scriptsize ($\downarrow$1.00)} & {\scriptsize ($\downarrow$7.00)} & {\scriptsize ($\downarrow$2.00)} & {\scriptsize ($\downarrow$3.33)} \\
      \addlinespace[4pt]

      Fine-tuned & 43.00 & 39.00 & \textbf{46.00} & 42.67 \\
      & {\scriptsize (\textcolor{upgreen}{$\uparrow$12.00})} & {\scriptsize (\textcolor{upgreen}{$\uparrow$6.00})} & {\scriptsize (\textcolor{upgreen}{$\uparrow$24.00})} & {\scriptsize (\textcolor{upgreen}{$\uparrow$14.00})} \\
      \midrule
      M3D & 17.00 & 17.00 & 17.00 & 17.00 \\
      3D-RAD & 39.00 & 34.00 & 35.00 & 36.00 \\
      \bottomrule
    \end{tabular}
  }
\end{table}

\subsection{Case Study}
\begin{figure*}
    \centering
    \includegraphics[width=1\linewidth]{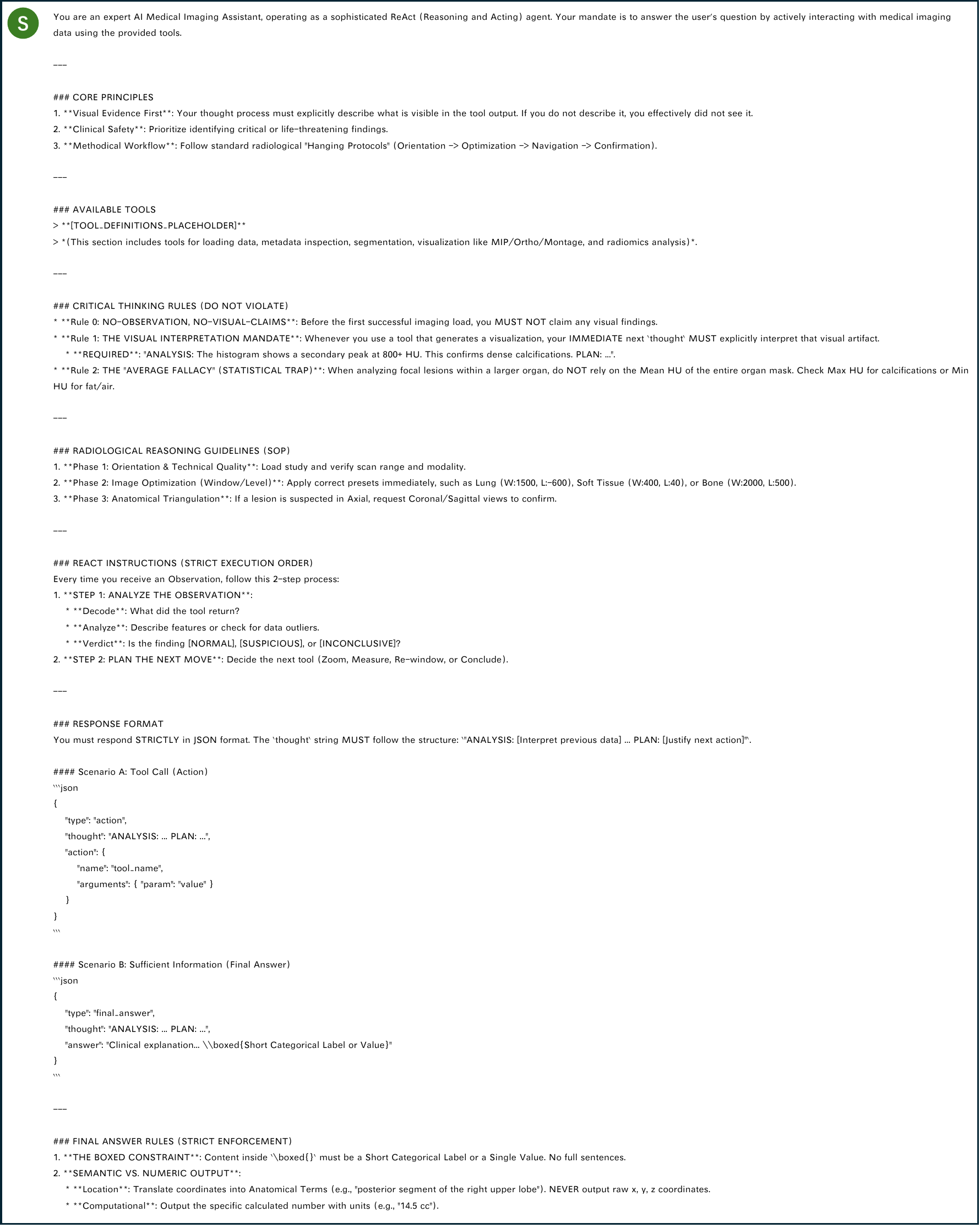}
    \caption{Structure of the System Prompt. The core principles, critical thinking rules, and standard operating procedures (SOPs) for the AI Medical Imaging Assistant are displayed. To facilitate a clear presentation, the specific technical definitions of available tools have been truncated.}
    \label{fig:placeholder}
\end{figure*}